\documentclass[11pt,a4paper]{article}
\usepackage[hyperref]{acl2018}
\usepackage{times}
\usepackage{latexsym}
\usepackage{amsmath}
\usepackage{amsfonts}
\usepackage[utf8]{inputenc}
\usepackage{amssymb}
\usepackage{multirow}
\usepackage{booktabs}
\usepackage{graphicx}
\usepackage{url}
\usepackage{subcaption}
\usepackage{comment}
\aclfinalcopy 
\def\aclpaperid{987}

\title{Reliability and Learnability of Human Bandit Feedback 
  \\ for Sequence-to-Sequence Reinforcement Learning}

\author{Julia Kreutzer$^{1}$ \and Joshua Uyheng$^{3}$\thanks{\;The work for this paper was done while the second author was an intern in Heidelberg.} \and Stefan Riezler$^{1,2}$ \\
$^1$Computational Linguistics \& $^2$IWR, Heidelberg University, Germany \\
{\tt \small \{kreutzer,riezler\}@cl.uni-heidelberg.de} \\
$^3$Departments of Psychology \& Mathematics, Ateneo de Manila University, Philippines\\
{\tt \small juyheng@ateneo.edu}
}

\date{}

\begin{document}
\maketitle

\begin{abstract}  
  We present a study on reinforcement learning (RL) from human bandit feedback for sequence-to-sequence learning, exemplified by the task of bandit neural machine translation (NMT). We investigate the reliability of human bandit feedback, and analyze the influence of reliability on the learnability of a reward estimator, and the effect of the quality of reward estimates on the overall RL task. Our analysis of cardinal (5-point ratings) and ordinal (pairwise preferences) feedback shows that their intra- and inter-annotator $\alpha$-agreement is comparable. Best reliability is obtained for standardized cardinal feedback, and cardinal feedback is also easiest to learn and generalize from. Finally, improvements of over 1 BLEU can be obtained by integrating a regression-based reward estimator trained on cardinal feedback for 800 translations into RL for NMT. This shows that RL is possible even from small amounts of fairly reliable human feedback, pointing to a great potential for applications at larger scale.
\end{abstract}

\section{Introduction}
\label{sec:intro}
Recent work has received high attention by successfully scaling reinforcement learning (RL) to games with large state-action spaces, achieving human-level \cite{MnihETAL:15} or even super-human performance \cite{SilverETAL:16}. This success and the ability of RL to circumvent the data annotation bottleneck in supervised learning has led to renewed interest in RL in sequence-to-sequence learning problems with exponential output spaces. A typical approach is to combine REINFORCE \cite{Williams:92} with policies based on deep sequence-to-sequence learning \cite{BahdanauETAL:15}, for example, in machine translation \cite{BahdanauETAL:17}, semantic parsing \cite{LiangETAL:17}, or summarization \cite{PaulusETAL:17}. These RL approaches focus on improving performance in automatic evaluation by simulating reward signals by evaluation metrics such as BLEU, F1-score, or ROUGE, computed against gold standards.
Despite coming from different fields of application, RL in games and sequence-to-sequence learning share
firstly the existence of a clearly specified reward function, e.g., defined by winning or losing a game, or by computing an automatic sequence-level evaluation metric. Secondly, both RL applications rely on a sufficient exploration of the action space, e.g., by evaluating multiple game moves for the same game state, or various sequence predictions for the same input.

The goal of this paper is to advance the state-of-the-art of sequence-to-sequence RL, exemplified by bandit learning for neural machine translation (NMT).  Our aim is to show that successful learning from simulated bandit feedback \cite{SokolovETALnips:16,KreutzerETAL:17,NguyenETAL:17,LawrenceETAL:17} does in fact carry over to learning from actual human bandit feedback. The promise of bandit NMT is that human feedback on the quality of translations is easier to obtain in large amounts than human references, thus compensating the weaker nature of the signals by their quantity. However, the human factor entails several differences to the above sketched simulation scenarios of RL. Firstly, human rewards are not well-defined functions, but complex and inconsistent signals. For example, in general every input sentence has a multitude of correct translations, each of which humans may judge differently, depending on many contextual and personal factors. Secondly, exploration of the space of possible  translations is restricted in real-world scenarios where a user judges one displayed translation, but cannot be expected to rate an alternative translation, let alone large amounts of alternatives.

In this paper we will show that despite the fact that human feedback is ambiguous and partial in nature, a catalyst for successful learning from human reinforcements is the reliability of the feedback signals. The first deployment of bandit NMT in an e-commerce translation scenario conjectured lacking reliability of user judgments as the reason for disappointing results when learning from 148k user-generated 5-star ratings for around 70k product title translations \cite{KreutzerETAL:18}. We thus raise the question of how human feedback can be gathered in the most reliable way, and what effect reliability will have in downstream tasks. In order to answer these questions, we measure intra- and inter-annotator agreement for two feedback tasks for bandit NMT, using cardinal feedback (on a 5-point scale) and ordinal feedback (by pairwise preferences) for 800 translations, conducted by 16 and 14 human raters, respectively. Perhaps surprisingly, while relative feedback is often considered easier for humans to provide \cite{Thurstone:27},
our investigation shows that $\alpha$-reliability \cite{Krippendorff:13} for intra- and inter-rater agreement is similar for both tasks, with highest inter-rater reliability for standardized 5-point ratings.

In a next step, we address the issue of machine learnability of human rewards. We use deep learning models to train reward estimators by regression against cardinal feedback, and by fitting a Bradley-Terry model \cite{BradleyTerry:52} to ordinal feedback. Learnability is understood by a slight misuse of the machine learning notion of learnability \cite{ShalevShwartzETAL:10} as the question how well reward estimates can approximate human rewards. 
Our experiments reveal that rank correlation of reward estimates with TER against human references is higher for regression models trained on standardized cardinal rewards than for Bradley-Terry models trained on pairwise preferences. This emphasizes the influence of the reliability of human feedback signals on the quality of reward estimates learned from them.

Lastly, we investigate machine learnability of the overall NMT task, in the sense of \citet{GreenETAL:14} who posed the question of how well an MT system can be tuned on post-edits. We use an RL approach for tuning, where a crucial difference of our work to previous work on RL from human rewards \cite{KnoxStone:09,ChristianoETAL:17} is that our RL scenario is not interactive, but rewards are collected in an offline log. RL then can proceed either by off-policy learning using logged single-shot human rewards directly, or by using estimated rewards. An expected advantage of estimating rewards is to tackle a simpler problem first --- learning a reward estimator instead of a full RL task for improving NMT --- and then to deploy unlimited feedback from the reward estimator for off-policy RL. Our results show that significant improvements can be achieved by training NMT from both estimated and logged human rewards, with best results for integrating a regression-based reward estimator into RL. This completes the argumentation that high reliability influences quality of reward estimates, which in turn affects the quality of the overall NMT task. Since the size of our training data is tiny in machine translation proportions, this result points towards a great potential for larger-scaler applications of RL from human feedback.

\section{Related Work}
\label{sec:related}
Function approximation to learn a ``critic'' instead of using rewards directly has been embraced in the RL literature under the name of ``actor-critic'' methods (see \citet{KondaTsitsiklis:00}, \citet{SuttonETAL:00}, \citet{Kakade:02}, \citet{SchulmanETAL:15}, \citet{MnihETAL:16}, \emph{inter alia}).
In difference to our approach, actor-critic methods learn online while our approach estimates rewards in an offline fashion.
Offline methods in RL, with and without function approximation, have been presented under the name of ``off-policy'' or ``counterfactual'' learning (see \citet{PrecupETAL:00}, \citet{PrecupETAL:01}, \citet{BottouETAL:13}, \citet{SwaminathanJoachims:15}, \citet{SwaminathanJoachimsNIPS:15}, \citet{JiangLi:16}, \citet{ThomasBrunskill:16}, \emph{inter alia}).
Online actor-critic methods have been applied to sequence-to-sequence RL by \citet{BahdanauETAL:17} and \citet{NguyenETAL:17}. An approach to off-policy RL under deterministic logging has been presented by \citet{LawrenceETAL:17}. However, all these approaches have been restricted to simulated rewards.

RL from human feedback is a growing area. \citet{KnoxStone:09} and \citet{ChristianoETAL:17} learn a reward function from human feedback and use that function to train an RL system. The actor-critic framework has been adapted to interactive RL from human feedback by \citet{PilarskiETAL:11} and \citet{MacGlashanETAL:17}. These approaches either update the reward function from human feedback intermittently or perform learning only in rounds where human feedback is provided. A framework that interpolates a human critique objective into RL has been presented by \citet{JudahETAL:10}. None of these works systematically investigates the reliability of the feedback and its impact of the down-stream task.

\citet{KreutzerETAL:18} have presented the first application of off-policy RL for learning from noisy human feedback obtained for deterministic logs of e-commerce product title translations. While learning from explicit feedback in the form of 5-star ratings fails, \citet{KreutzerETAL:18} propose to leverage implicit feedback embedded in a search task instead. In simulation experiments on the same domain, the methods proposed by \citet{LawrenceETAL:17} succeeded also for neural models, allowing to pinpoint the lack of reliability in the human feedback signal as the reason for the underwhelming results when learning from human 5-star ratings. The goal of showing the effect of highly reliable human bandit feedback in downstream RL tasks was one of the main motivations for our work.

For the task of machine translation, estimating human feedback, i.e. quality ratings, is related to the task of sentence-level quality estimation (sQE). However, there are crucial differences between sQE and the reward estimation in our work: sQE usually has more training data, often from more than one machine translation model. Its gold labels are inferred from post-edits, i.e. corrections of the machine translation output, while we learn from weaker bandit feedback. Although this would in principle be possible, sQE predictions have not (yet) been used to directly reinforce predictions of MT systems, mostly because their primary purpose is to predict post-editing effort, i.e. give guidance how to further process a translation.  
State-of-the-art models for sQE such as \cite{MartinsETAL:17} and \cite{KimETAL:17} are unsuitable for the direct use in this task since they rely on linguistic input features, stacked architectures or post-edit or word-level supervision.
Similar to approaches for generative adversarial NMT \cite{YuETAL:17, WuETAL:17} we prefer a simpler convolutional architecture based on word embeddings for the human reward estimation.

\section{Human MT Rating Task}
\label{sec:data}
\begin{figure}[t]
	\includegraphics[width=\columnwidth]{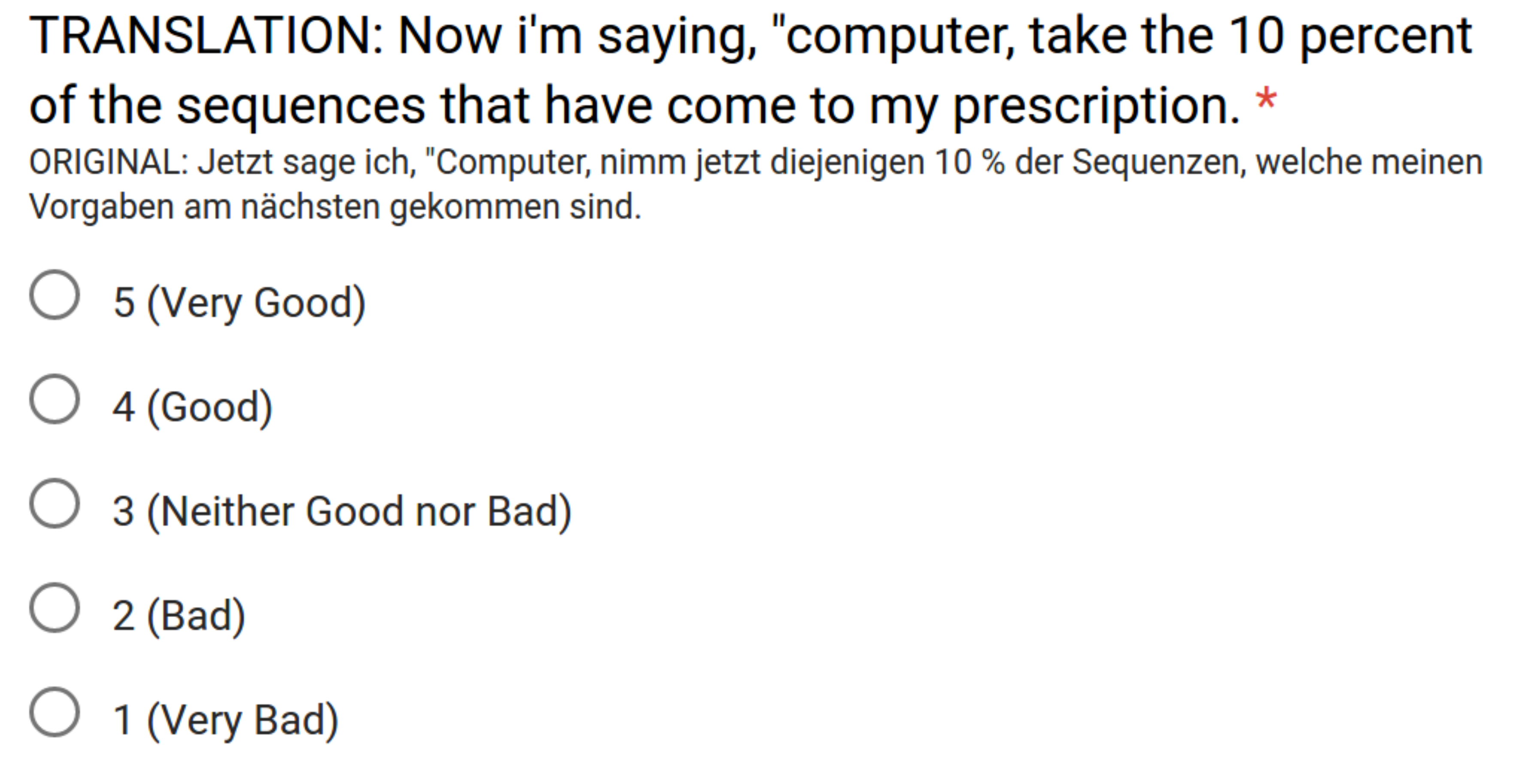}
	\caption{Rating interface for 5-point ratings.}
	\label{fig:rating-5}
\end{figure}

\begin{figure}[t]
	\includegraphics[width=\columnwidth]{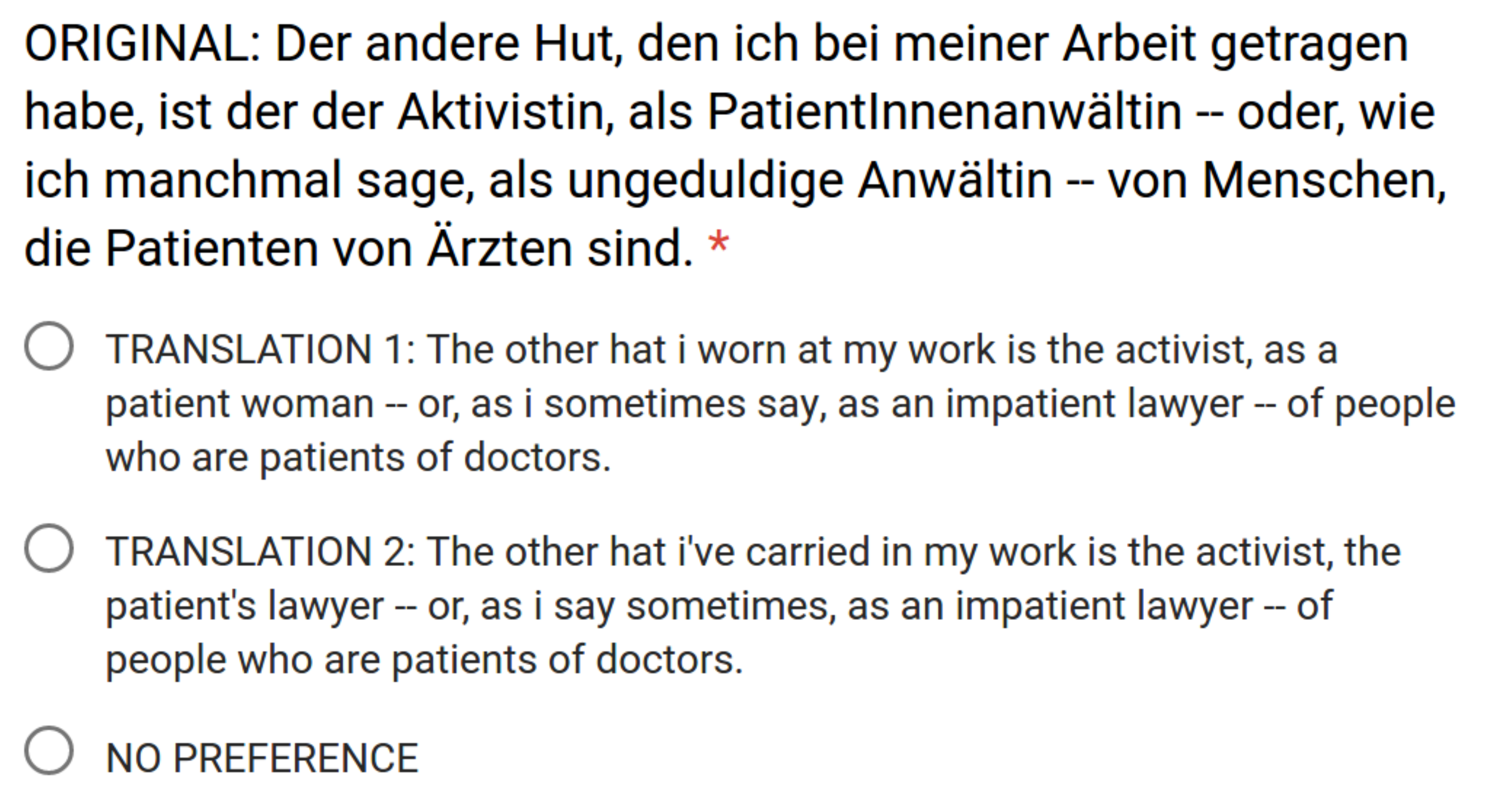}
	\caption{Rating interface for pairwise ratings.}
	\label{fig:rating-pw}
\end{figure}

\subsection{Data}
We translate a subset of the TED corpus with a general-domain and a domain-adapted NMT model (see §\ref{sec:rl-setup} for NMT and data), post-process the translations (replacing special characters, restoring capitalization) and filter out identical out-of-domain and in-domain translations. In order to compose a homogeneous data set, we first select translations with references of length 20 to 40, then sort the translation pairs by difference in character n-gram F-score (chrF, $\beta=3$) \cite{Popovic:2015:WMT} and length, and pick the top 400 translation pairs with the highest difference in chrF but lowest difference in length. This yields translation pairs of similar length, but different quality.

\subsection{Rating Task}
The pairs were treated as 800 separate translations for a 5-point rating task. From the original 400 translation pairs, 100 pairs (or 200 individual translations) were randomly selected for repetition. This produced a total of 1,000 individual translations, with 600 occurring once, and 200 occurring twice. The translations were shuffled and separated into five sections of 200 translations, each with 120 translations from the unrepeated pool, and 80 translations from the repeated pool, ensuring that a single translation does not occur more than once in each section. 
For a pairwise task, the same 100 pairs were repeated from the original 400 translation pairs. This produced a total of 500 translation pairs. The translations were also shuffled and separated into five sections of 100 translation pairs, each with 60 translation pairs from the unrepeated pool, and 40 translation pairs from the repeated pool. None of the pairs were repeated within each section. 

We recruited 14 participants for the pairwise rating task and 16 for the 5-point rating task. The participants were university students with fluent or native language skills in German and English. The rating interface is shown in Figures \ref{fig:rating-5} and \ref{fig:rating-pw}. Rating instructions are given in Appendix \ref{app:ratings}. Note that no reference translations were presented since the objective is to model a realistic scenario for bandit learning.\footnote{The collection of ratings can be downloaded from \url{http://www.cl.uni-heidelberg.de/statnlpgroup/humanmt/}.}

\section{Reliability of Human MT Ratings}
\label{sec:reliability}
\subsection{Inter-rater and Intra-rater Reliability}

\begin{table}[t]
\resizebox{\columnwidth}{!}{
\begin{tabular}{l|c|cc}
\toprule
 & \textbf{Inter-rater} & \multicolumn{2}{c}{\textbf{Intra-rater}}  \\
\textbf{Type} & $\mathbf{\alpha}$ & \textbf{Mean} $\mathbf{\alpha}$ & \textbf{Stdev.} $\mathbf{\alpha}$ \\
\midrule
5-point & 0.2308 & \multirow{2}{*}{0.4014} & \multirow{2}{*}{0.1907} \\
5-point norm. & 0.2820 & & \\
\midrule
5-point norm. part.& 0.5059 & 0.5527 & 0.0470 \\
5-point norm. trans. & 0.3236 & 0.3845 & 0.1545\\
\bottomrule
Pairwise & 0.2385 & 0.5085 & 0.2096 \\
\midrule
Pairwise filt. part. & 0.3912 & 0.7264 & 0.0533 \\
Pairwise filt. trans.& 0.3519 & 0.5718 & 0.2591 \\
\bottomrule
\end{tabular}%
}
\caption{Inter- and intra-reliability measured by Krippendorff's $\alpha$ for 5-point and pairwise ratings of 1,000 translations of which 200 translations are repeated twice. The filtered variants are restricted to either a 	subset of participants (part.) or a subset of translations (trans.).}
\label{tab:reliability}
\end{table}

In the following, we report inter- and intra-rater reliability of the cardinal and ordinal feedback tasks described in §\ref{sec:data} with respect to Krippendorff's $\alpha$ \cite{Krippendorff:13} evaluated at interval and ordinal scale, respectively.

As shown in Table \ref{tab:reliability}, measures of inter-rater reliability show small differences between the 5-point and pairwise task. The inter-rater reliability in the 5-point task $\left(\alpha=0.2308\right)$ is roughly the same as that of the pairwise task $\left(\alpha=0.2385\right)$. Normalization of ratings per participant (by standardization to Z-scores), however, shows a marked improvement of overall inter-rater reliability for the 5-point task $\left(\alpha=0.2820\right)$. A one-way analysis of variance taken over inter-rater reliabilities between pairs of participants suggests statistically significant differences across tasks $\left(F\left(2,328\right)=6.399,p<0.01\right)$, however, a post hoc Tukey's \cite{LarsenMarx:12} honest significance test attributes statistically significant differences solely between the 5-point tasks with and without normalization. 
These scores indicate that the overall agreement between human ratings is roughly the same, regardless of whether participants are being asked to provide cardinal or ordinal ratings. Improvement in inter-rater reliability via participant-level normalization suggests that participants may indeed have individual biases toward certain regions of the 5-point scale, which the normalization process corrects. 

In terms of intra-rater reliability, a better mean was observed among participants in the pairwise task $\left(\alpha=0.5085\right)$ versus the 5-point task $\left(\alpha=0.4014\right)$. This suggests that, on average, human raters provide more consistent ratings with themselves in comparing between two translations versus rating single translations in isolation. This may be attributed to the fact that seeing multiple translations provides raters with more cues with which to make consistent judgments. 
However, at the current sample size, a Welch two-sample t-test \cite{LarsenMarx:12} between 5-point 
and pairwise 
intra-rater reliabilities shows no significant difference between the two tasks $\left(t\left(26.92\right)=1.4362,p=0.1625\right)$. Thus, it remains difficult to infer whether one task is definitively superior to the other in eliciting more consistent responses. 
Intra-rater reliability is the same for the 5-point task with and without normalization, as participants are still compared against themselves.

\subsection{Rater and Item Variance}
The succeeding analysis is based on two assumptions: first, that human raters vary in that they do not provide equally good judgments of translation quality, and second, rating items vary in that some translations may be more difficult to judge than others. This allows to investigate the influence of rater variance and item variance on inter-rater reliability by an ablation analysis where low-quality judges and difficult translations are filtered out.

\begin{figure}[t]
	\includegraphics[width=0.9\columnwidth]{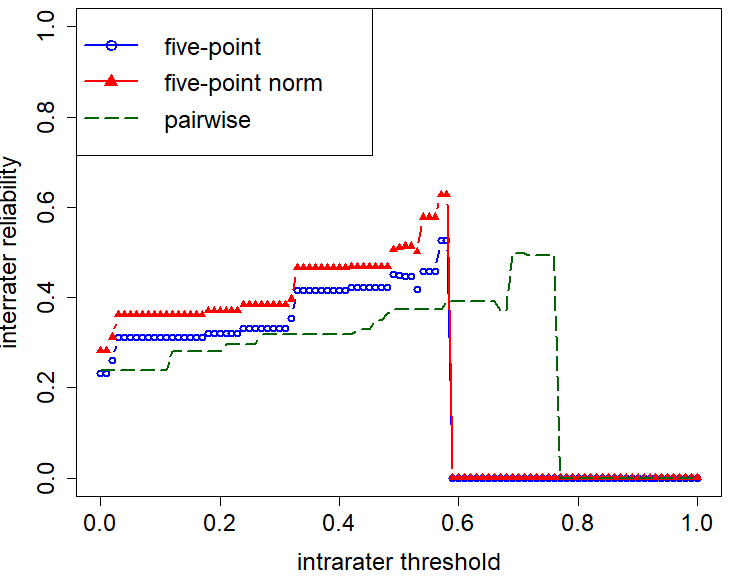}
	\caption{Improvements in inter-rater reliability using \emph{intra-rater consistency} filter.}
	\label{fig:consistencyfilter}
\end{figure}

\begin{figure}[t]
	\includegraphics[width=0.9\columnwidth]{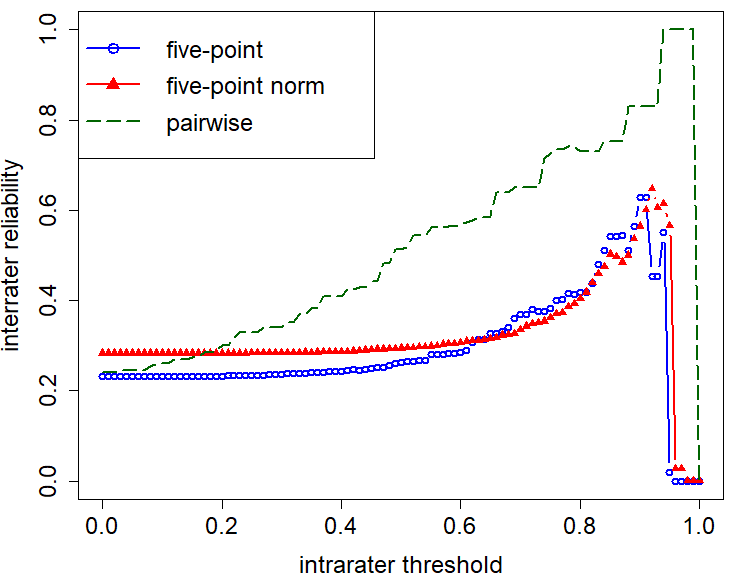}
	\caption{Improvements in inter-rater reliability using \emph{item variance} filter.}
	\label{fig:variancefilter}
\end{figure}

Using intra-rater reliability as an index of how well human raters judge translation quality, Figure \ref{fig:consistencyfilter} shows a filtering process whereby human raters with $\alpha$ scores lower than a moving threshold are dropped from the analysis. As the reliability threshold is increased from $0$ to $1$, overall inter-rater reliability is measured. Figure \ref{fig:variancefilter} shows a similar filtering process implemented using variance in translation scores. Item variances are normalized on a scale from $0$ to $1$ and subtracted from $1$ to produce an item variance threshold. As the threshold increases, overall inter-rater reliability is likewise measured as high-variance items are progressively dropped from the analysis.

As the plots demonstrate, inter-rater reliability generally increases with consistency and variance filtering. For consistency filtering, Figure \ref{fig:consistencyfilter} shows how the inter-rater reliability of the 5-point task experiences greater increases than the pairwise task with lower filtering thresholds, especially in the normalized case. This may be attributed to the fact that more participants in the 5-point task had low intra-rater reliability. Pairwise tasks, on the other hand, require higher thresholds before large gains are observed in overall inter-rater reliability. This is because more participants in the pairwise task had relatively high intra-rater reliability.
In the normalized 5-point task, selecting a threshold of 0.49 as a cutoff for intra-rater reliability retains 8 participants with an inter-rater reliability of 0.5059. For the pairwise task, a threshold of 0.66 leaves 5 participants with an inter-rater reliability of 0.3912.

The opposite phenomenon is observed in the case of variance filtering. As seen in Figure \ref{fig:variancefilter}, the overall inter-rater reliability of the pairwise task quickly overtakes that of the 5-point task, with and without normalization. This may be attributed to how, in the pairwise setup, more items can be a source of disagreement among human judges. Ambiguous cases, that will be discussed in §\ref{sec:qual},
may result in higher item variance.  This problem is not as pronounced in the 5-point task, where judges must simply rate individual translations. It may be surmised that this item variance accounts for why, on average, judges in the pairwise task demonstrate higher intra-rater reliability than those in the 5-point task, yet the overall inter-rater reliability of the pairwise task is lower. 
By selecting a variance threshold such that at least $70\%$ of items are retained in the analysis, the improved inter-rater reliabilities were 0.3236 for the 5-point task and 0.3519 for the pairwise task.

\subsection{Qualitative Analysis}\label{sec:qual}
On completion of the rating task, we asked the participants for a \emph{subjective} judgment of difficulty on a scale from 1 (very difficult) to 10 (very easy). On average, the pairwise rating task (mean 5.69) was perceived slightly easier than the 5-point rating task (mean 4.8). They also had to state which aspects of the tasks they found difficult: The biggest challenge for 5-point ratings seemed to be the weighing of different error types and the rating of long sentences with very few, but essential errors. For pairwise ratings, difficulties lie in distinguishing between similar, or similarly bad translations. Both tasks showed difficulties with ungrammatical or incomprehensible sources.

Comparing items with high and low agreement across raters allows conclusions about \emph{objective} difficulty. We assume that high inter-rater agreement indicates an ease of judgment, while difficulties in judgment are manifested in low agreement. A list of examples is given in Appendix \ref{app:examples}.
For 5-point ratings, difficulties arise with ungrammatical sources and omissions, whereas obvious mistakes in the target, such as over-literal translations, make judgment easier. Preference judgments tend to be harder when both translations contain errors and are similar. When there is a tie, the pairwise rating framework does not allow to indicate whether both translations are of high or low quality. Since there is no normalization strategy for pairwise ratings, individual biases or rating schemes can hence have a larger negative impact on the inter-rater agreement.

\section{Learnability of a Reward Estimator from MT Ratings}
\label{sec:learnability}
\subsection{Learning a Reward Estimator}
The numbers of ratings that can be obtained directly from human raters in a reasonable amount of time is tiny compared to the millions of sentences used for standard NMT training. By learning a reward estimator on the collection of human ratings, we seek to generalize to unseen translations. The model for this reward estimator should ideally work without time-consuming feature extraction so it can be deployed in direct interaction with a learning NMT system, estimating rewards on the fly, and most importantly generalize well so it can guide the NMT towards good local optima.

\paragraph{Learning from Cardinal Feedback.}
The inputs to the reward estimation model are sources $\mathbf{x}$ and their translations $\mathbf{y}$. Given cardinal judgments for these inputs, a regression model with parameters $\boldsymbol \psi$ is trained to minimize the mean squared error (MSE) for a set of $n$ predicted rewards $\hat{r}$ and judgments $r$:
\begin{align*}
\mathcal{L}^{MSE}(\boldsymbol \psi) = \frac{1}{n}\sum^n_{i=1}(r(\mathbf{y}_i) -\hat{r}_{\boldsymbol \psi}(\mathbf{y}_i))^2.
\end{align*}
In simulation experiments, where all translations can be compared to existing references, $r$ may be computed by sentence-BLEU (sBLEU). For our human 5-point judgments, we first normalize the judgments per rater as described in §\ref{sec:reliability}, then average the judgments across raters and finally scale them linearly to the interval $[0.0, 1.0]$.

\paragraph{Learning from Pairwise Preference Feedback.}
When pairwise preferences are given instead of cardinal judgments, the Bradley-Terry model allows us to train an estimator of $r$. Following \citet{ChristianoETAL:17}, let $\hat{P}_{\boldsymbol \psi}[\mathbf{y^1} \succ \mathbf{y^2}]$ be the probability that any translation $\mathbf{y^1}$ is preferred over any other translation $\mathbf{y^2}$ by the reward estimator:
\begin{align*}
\hat{P}_{\boldsymbol \psi}[\mathbf{y}^1 \succ \mathbf{y}^2] = \frac{\exp \hat{r}_{\boldsymbol \psi}(\mathbf{y^1})}{\exp \hat{r}_{\boldsymbol \psi}(\mathbf{y^1}) + \exp \hat{r}_{\boldsymbol \psi}(\mathbf{y^2})}.
\end{align*}
Let $Q[\mathbf{y^1} \succ \mathbf{y^2}]$ be the probability that translation $\mathbf{y_1}$ is preferred over translation $\mathbf{y_2}$ by a gold standard, e.g. the human raters or in comparison to a reference translation. With this supervision signal we formulate a pairwise (PW) training loss for the reward estimation model with parameters $\boldsymbol \psi$:
\begin{align*}
\mathcal{L}^{PW}(\boldsymbol \psi) = -\frac{1}{n}\sum^n_{i=1}&Q[\mathbf{y^1_i} \succ \mathbf{y^2_i}] \log \hat{P}_{\boldsymbol \psi}[\mathbf{y^1_i} \succ \mathbf{y^2_i}] \nonumber\\ 
+ &Q[\mathbf{y^2_i} \succ \mathbf{y^1_i}] \log \hat{P}_{\boldsymbol \psi}[\mathbf{y^2_i} \succ \mathbf{y^1_i}].
\end{align*}
For simulation experiments --- where we lack a genuine supervision for preferences --- we compute $Q$ comparing the sBLEU scores for both translations, i.e. translation preferences are modeled according to their difference in sBLEU:
\begin{align*}
Q[\mathbf{y^1} \succ \mathbf{y^2}] = \frac{\exp \text{sBLEU}(\mathbf{y^1})}{\exp \text{sBLEU}(\mathbf{y^1}) + \exp \text{sBLEU}(\mathbf{y^2})}.
\end{align*}
When obtaining preference jugdments directly from raters, $Q[\mathbf{y^1} \succ \mathbf{y^2}]$ is simply the relative frequency of $\mathbf{y^1}$ being preferred over $\mathbf{y^2}$ by a rater.

\subsection{Experiments}

\paragraph{Data.}
The 1,000 ratings collected as described in §\ref{sec:data} are leveraged to train regression models and pairwise preference models. In addition, we train models on simulated rewards (sBLEU) for a comparison with arguably ``clean'' feedback for the same set of translations. 
In order to augment this very small collection of ratings, we leverage the available out-of-domain bitext as auxiliary training data. We sample translations for a subset of the out-of-domain sources and store sBLEU scores as rewards, collecting 90k out-of-domain training samples in total (see Appendix \ref{app:re} for details). During training, each mini-batch is sampled from the auxiliary data with probability $p_{aux}$, from the original training data with probability $1-p_{aux}$. Adding this auxiliary data as a regularization through multi-task learning prevents the model from overfitting to the small set of human ratings. In the experiments $p_{aux}$ was tuned to 0.8.

\paragraph{Architecture.}
We choose the following neural architecture for the reward estimation (details in Appendix \ref{app:nn}): Inputs are padded source and target subword embeddings, which are each processed with a biLSTM \cite{HochreiterSchmidhuber:97}. Their outputs are concatenated for each time step, further fed to a 1D-convolution with max-over-time pooling and subsequently a leaky ReLU \cite{MaasETAL:13} output layer. This architecture can be seen as a biLSTM-enhanced bilingual extension to the convolutional model for sentence classification proposed by \citet{Kim:14}. It has the advantage of not requiring any feature extraction but still models n-gram features on an abstract level.
 
\paragraph{Evaluation Method.}
The quality of the reward estimation models is tested by measuring Spearman's $\rho$ with TER on a held-out test set of 1,314 translations following the standard in sQE evaluations. Hyperparameters are tuned on another 1,200 TED translations. 

\paragraph{Results.}\label{sec:estimation-results}

\begin{table}
\center
\begin{tabular}{llc}
\toprule
\textbf{Model} & \textbf{Feedback} & $\rho$ \\
\midrule
MSE & Simulated & -0.2571 \\ 
PW & Simulated & -0.1307 \\ 
\midrule
MSE & Human & -0.2193 \\ 
PW & Human & -0.1310 \\ 
\midrule
MSE & Human filt.& -0.2341\\
PW & Human filt.& -0.1255 \\
\bottomrule
\end{tabular} 
\caption{Spearman's rank correlation $\rho$ between estimated rewards and TER for models trained with \emph{simulated} rewards and \emph{human} rewards (also filtered subsets).}
\label{tab:estimations}
\end{table}

Table \ref{tab:estimations} reports the results of reward estimators trained on simulated and human rewards. When trained from cardinal rewards, the model of simulated scores performs slightly better than the model of human ratings. This advantage is lost when moving to preference judgments, which might be explained by the fact that the softmax over sBLEUs with respect to a single reference is just not as expressive as the preference probabilities obtained from several raters. Filtering by participants (retaining 8 participants for cardinal rewards and 5 for preference jugdments, see Section \ref{sec:reliability}) improves the correlation further for cardinal rewards, but slightly hurts for preference judgments. The overall correlation scores are relatively low --- especially for the PW models --- which we suspect is due to overfitting to the small set of training data. From these experiments we conclude that when it comes to estimating translation quality, cardinal human jugdments are more useful than pairwise preference jugdments.

\section{Reinforcement Learning from Direct and Estimated Rewards in MT}
\label{sec:rl}
\subsection{NMT Objectives}

\paragraph{Supervised Learning.} Most commonly, NMT models are trained with Maximum Likelihood Estimation (MLE) on a parallel corpus of source and target sequences $D = \{(\mathbf{x}^{(s)}, \mathbf{y}^{(s)})\}^{S}_{s=1}$:
\begin{align*}
\mathcal{L}^{MLE}(\boldsymbol \theta) = \sum_{s=1}^S \log p_{\theta}(\mathbf{y}^{(s)}|\mathbf{x}^{(s)}). 
\end{align*} 
The MLE objective requires reference translations and is agnostic to rewards. In the experiments it is used to train the out-of-domain baseline model as a warm start for reinforcement learning from in-domain rewards.

\paragraph{Reinforcement Learning from Estimated or Simulated Direct Rewards.} 

Deploying NMT in a reinforcement learning scenario, the goal is to maximize 
the expectation of a reward $r$ over all source and target sequences \cite{WuETAL:16}, leading to the following REINFORCE \cite{Williams:92} objective:
\begin{align} 
\mathcal{R}^{RL}(\boldsymbol \theta) =& \mathbb{E}_{p(\mathbf{x}) p_{\boldsymbol \theta}(\mathbf{y}|\mathbf{x})} \left[r(\mathbf{y})\right] \label{eq:rl}\\ 
\approx& \sum_{s=1}^S \sum_{i=1}^{k} p^{\tau}_{\boldsymbol \theta}(\mathbf{\tilde{y}^{(s)}_i}|\mathbf{x^{(s)}})\, r(\mathbf{\tilde{y}_i}) \label{eq:approx-rl}
\end{align}
The reward $r$ can either come from a reward estimation model (\emph{estimated reward}) or be computed with respect to a reference in a simulation setting (\emph{simulated direct reward}). In order to counteract high variance in the gradient updates, the running average of rewards is subtracted from $r$ for learning.
In practice, Equation \ref{eq:rl} is approximated with $k$ samples from $p_{\boldsymbol \theta}(\mathbf{y}|\mathbf{x})$ (see Equation \ref{eq:approx-rl}). When $k=1$, this is equivalent to the expected loss minimization in \citet{SokolovETAL:16, SokolovETALnips:16, KreutzerETAL:17}, where the system interactively learns from online bandit feedback. For $k>1$ this is similar to the minimum-risk training for NMT proposed in \citet{ShenETAL:16}. 
Adding a temperature hyper-parameter $\tau \in (0.0, \infty]$ to the softmax over the model output $\mathbf{o}$ allows us to control the sharpness of the sampling distribution $p^{\tau}_{\boldsymbol \theta}(\mathbf{y}|\mathbf{x}) = \text{softmax}(\mathbf{o}/\tau)$, i.e. the amount of exploration during training.
With temperature $\tau<1$, the model's entropy decreases and samples closer to the one-best output are drawn. We seek to keep the exploration low to prevent the NMT to produce samples that lie far outside the training domain of the reward estimator.

\paragraph{Off-Policy Learning from Direct Rewards.} 
When rewards can not be obtained for samples from a learning system, but were collected for a static deterministic system (e.g. in a production environment), we are in an \emph{off-policy learning} scenario. 
The challenge is to improve the MT system from a log $L = \{(\mathbf{x}^{(h)}, \mathbf{y}^{(h)}, r(\mathbf{y}^{(h)}))\} ^H_{h=1}$ of rewarded translations.
Following \citet{LawrenceETAL:17} we define the following off-policy learning (OPL) objective to learn from logged rewards:
\begin{align*} \label{eq:opl-loss}
\mathcal{R}^{OPL}(\boldsymbol \theta) = \frac{1}{H}\sum_{h=1}^H r(\mathbf{y}^{(h)}) \, \bar{p}_{\boldsymbol \theta}(\mathbf{y}^{(h)}|\mathbf{x}^{(h)}),
\end{align*} 
with reweighting over the current mini-batch $B$: $\bar{p}_{\boldsymbol \theta}(\mathbf{y}^{(h)}|\mathbf{x}^{(h)}) =
\frac{p_{\boldsymbol \theta}(\mathbf{y}^{(h)}|\mathbf{x}^{(h)})}{\sum_{b=1}^B p_{\boldsymbol \theta}(\mathbf{y}^{(b)}|\mathbf{x}^{(b)})}$.\footnote{\citet{LawrenceETAL:17} propose reweighting over the whole log, but this is infeasible for NMT. Here $B \ll H$.} In contrast to the RL objective, only logged translations are reinforced, i.e. there is no exploration in learning.

\subsection{Experiments}\label{sec:rl-setup}

\paragraph{Data.}
We use the WMT 2017 data\footnote{Pre-processed data available at \url{http://www.statmt.org/wmt17/translation-task.html}.} for training a general domain (here: \emph{out-of-domain}) model for translations from German to English. The training data contains 5.9M sentence pairs, the development data 2,999 sentences (WMT 2016 test set) and the test data 3,004 sentences.
For \emph{in-domain} data, we choose the translations of TED talks\footnote{Pre-processing and data splits as described in \url{https://github.com/rizar/actor-critic-public/tree/master/exp/ted}.} as used in IWSLT evaluation campaigns. The training data contains 153k, the development data 6,969, and the test data 6,750 parallel sentences.

\paragraph{Architecture.} Our NMT model is a standard subword-based encoder-decoder architecture with attention \cite{BahdanauETAL:15}. An encoder Recurrent Neural Network (RNN) reads in the source sentence and a decoder RNN generates the target sentence conditioned on the encoded source. We implemented RL and OPL objectives in Neural Monkey \cite{NeuralMonkey:2017}.\footnote{The code is available in the Neural Monkey fork \url{https://github.com/juliakreutzer/bandit-neuralmonkey/tree/acl2018}.} The NMT has a bidirectional encoder and a single-layer decoder with 1,024 GRUs each, and subword embeddings of size 500 for a shared vocabulary of subwords obtained from 30k byte-pair merges \cite{SennrichETAL:16}. For model selection we use greedy decoding, for test set evaluation beam search with a beam of width 10. 
We sample $k=5$ translations for RL models and set the softmax temperature $\tau=0.5$. Appendix \ref{app:hyper} reports remaining hyperparameters.

\paragraph{Evaluation Method.}
Trained models are evaluated with respect to BLEU \cite{PapineniETAL:02}, METEOR \cite{DenkowskiLavie:11} using MULTEVAL \cite{ClarkETAL:11} and BEER  \cite{StanojevicSimaan:14} to cover a diverse set of automatic measures for translation quality.\footnote{Since tendencies of improvement turn out to be consistent across metrics, we only discuss BLEU in the text.}
We test for statistical significance with approximate randomization \cite{Noreen:89}.

\begin{table}
\center
\resizebox{\columnwidth}{!}{
\begin{tabular}{l|ccc|ccc}
\toprule
& \multicolumn{3}{c|}{\textbf{WMT}} & \multicolumn{3}{c}{\textbf{TED}} \\
\textbf{Model} & BLEU & METEOR & BEER & BLEU & METEOR & BEER\\
\midrule
WMT & 27.2 & 31.8  & 60.08 & 27.0 & 30.7  & 59.48\\
TED & 26.3 & 31.3  & 59.49 & 34.3 & 34.6  & 64.94\\
\bottomrule
\end{tabular} %
}
\caption{Results on test data for in- and out-of-domain \emph{fully-supervised} models. Both are trained with MLE, the TED model is obtained by fine-tuning the WMT model on TED data.}
\label{tab:sup}
\end{table}

The out-of-domain model is trained with MLE on WMT. The task is now to improve the generalization of this model to the TED domain. Table \ref{tab:sup} compares the out-of-domain baseline with domain-adapted models that were further trained on TED in a fully-supervised manner (\emph{supervised fine-tuning} as introduced by \citet{FreitagETAL:16, LuongManning:15}). The supervised domain-adapted model serves as an upper bound for domain adaptation with human rewards: if we had references, we could improve up to 7 BLEU. What if references are not available, but we can obtain rewards for sample translations?  

\begin{table}
\center
\resizebox{\columnwidth}{!}{
\begin{tabular}{lll|ccc}
\toprule
\textbf{Model} & \multicolumn{2}{c|}{\textbf{Rewards}} & \textbf{BLEU} & \textbf{METEOR} & \textbf{BEER} \\
\midrule
Baseline & - & - & 27.0 & 30.7 & 59.48 \\
\midrule
\midrule
RL & D & S & 32.5$_{\pm 0.01}^{\star}$ & 33.7$_{\pm 0.01}^{\star}$ & 63.47$_{\pm 0.10}^{\star}$ \\
OPL & D & S & 27.5$^{\star}$ & 30.9$^{\star}$  & 59.62$^{\star}$\\
\midrule
RL+MSE & E & S & 28.2$_{\pm 0.09}^{\star}$ & 31.6$_{\pm 0.04}^{\star}$ & 60.23$_{\pm 0.14}^{\star}$ \\ 
RL+PW & E & S & 27.8$_{\pm 0.01}^{\star}$ & 31.2$_{\pm 0.01}^{\star}$ & 59.83$_{\pm 0.04}^{\star}$ \\ 
\midrule
\midrule
OPL & D & H & 27.5$^{\star}$ & 30.9$^{\star}$ & 59.72$^{\star}$\\
\midrule
RL+MSE & E & H & 28.1$_{\pm 0.01}^{\star}$ & 31.5$_{\pm 0.01}^{\star}$ & 60.21$_{\pm 0.12}^{\star}$\\ 
RL+PW & E & H & 27.8$_{\pm 0.09}^{\star}$ & 31.3$_{\pm 0.09}^{\star}$  & 59.88$_{\pm 0.23}^{\star}$\\ 
\midrule
RL+MSE & E & F & 28.1$_{\pm 0.20}^{\star}$ & 31.6$_{\pm 0.10}^{\star}$ & 60.29$_{\pm 0.13}^{\star}$\\ 
\bottomrule
\end{tabular} %
}
\caption{Results on TED test data for training with \emph{estimated} (E) and \emph{direct} (D) rewards from \emph{simulation} (S), \emph{humans} (H) and \emph{filtered} (F) human ratings. Significant ($p\leq0.05$) differences to the baseline are marked with $^{\star}$. For RL experiments we show three runs with different random seeds, mean and standard deviation in subscript.}
\label{tab:rl}
\end{table}

\paragraph{Results for RL from Simulated Rewards.} First we simulate ``clean'' and deterministic rewards by comparing sample translations to references using GLEU \cite{WuETAL:16} for RL, and smoothed sBLEU for estimated rewards and OPL. 
Table \ref{tab:rl} lists the results for this simulation experiment in rows 2-5 (S). 
If unlimited clean feedback was given (RL with direct 
simulated rewards), improvements of over 5 BLEU can be achieved. When limiting the amount of feedback to a log of 800 translations, the improvements over the baseline are only marginal (OPL). When replacing the direct reward by the simulated reward estimators from §\ref{sec:learnability}, i.e. having unlimited amounts of approximately clean rewards, however, improvements of 1.2 BLEU for MSE estimators (RL+MSE) and 0.8 BLEU for pairwise estimators (RL+PW) are found. This suggests that the reward estimation model helps to tackle the challenge of generalization over a small set of ratings.

\paragraph{Results for RL from Human Rewards.} Knowing what to expect in an ideal setting with non-noisy feedback, we now move to the experiments with human feedback. OPL is trained with the logged normalized, averaged and re-scaled human reward (see §\ref{sec:learnability}). RL is trained with the direct reward provided by the reward estimators trained on human rewards from §\ref{sec:learnability}. Table \ref{tab:rl} shows the results for training with human rewards in rows 6-8: The improvements for OPL are very similar to OPL with simulated rewards, both suffering from overfitting. For RL we observe that the MSE-based reward estimator (RL+MSE) leads to significantly higher improvements as a the pairwise reward estimator (RL+PW) --- the same trend as for simulated ratings. Finally, the improvement of 1.1 BLEU over the baseline showcases that we are able to improve NMT with only a small number of human rewards. Learning from estimated filtered 5-point ratings, does not significantly improve over these results, since the improvement of the reward estimator is only marginal (see § \ref{sec:learnability}).

\section{Conclusion}
\label{sec:disc}
In this work, we sought to find answers to the questions of how cardinal and ordinal feedback differ in terms of reliability, learnability and effectiveness for RL training of NMT, with the goal of improving NMT with human bandit feedback. Our rating study, comparing 5-point and preference ratings, showed that their reliability is comparable, whilst cardinal ratings are easier to learn and to generalize from, and also more suitable for RL in our experiments.

Our work reports improvements of NMT leveraging actual human bandit feedback for RL, leaving the safe harbor of simulations. Our experiments show that improvements of over 1 BLEU are achievable by learning from a dataset that is tiny in machine translation proportions. Since this type of feedback, in contrast to post-edits and references, is fast and cheap to elicit from non-professionals, our results bear a great potential for future applications on larger scale.

\section*{Acknowledgments.} This work was supported in part by DFG Research Grant RI 2221/4-1, and by an internship program of the IWR at Heidelberg University.

\bibliography{references}
\bibliographystyle{acl_natbib}

\clearpage

\section*{Appendix}
\appendix
\section{Rating Task}

\subsection{Rating Instructions}\label{app:ratings}
Participants for the 5-star rating task were given the following instructions: ``You will be presented with a German statement and a translation of this statement in English. You must assign a rating from 1 (Very Bad) to 5 (Very Good) to each translation.''

Participants for the pairwise task were given the following instructions: ``You will be presented with a German statement and two translations of this statement in English. You must decide which of the two translations you prefer, or whether you have no preference.''

\subsection{Example Ratings}\label{app:examples}
Table \ref{tab:5star-ex} lists low- and high-variance items for 5-star ratings, Table \ref{tab:pw-ex} for pairwise ratings. From the annotations in the tables, the reader may get an impression which translations are ``easier'' to judge than others.

\begin{table*}
\resizebox{\textwidth}{!}{
\begin{tabular}{ll|ll}
\toprule
\multirow{3}{*}{\#1} & source & Diese könnten Kurierdienste sein, oder Techniker zum Beispiel, nur um sicherzustellen, dass der gemeldete AED sich immer noch an seiner Stelle befindet.\\
& target & These could be courier services, or technicians like, for example, just to make sure that the \underline{abalone aed} is still in its place.\\
& rating & $\sigma=0.46$, $\varnothing=-0.30$\\
\midrule
\multirow{3}{*}{\#2} & source & Es muss für mich im Hier und Jetzt stimmig sein, sonst kann ich mein Publikum nicht davon überzeugen, dass das mein Anliegen ist.\\
& target & It must \underline{be for me here and now}, otherwise i cannot convince my audience that my concern is.\\
& rating & $\sigma=0.46$, $\varnothing=-0.70$\\
\midrule
\multirow{3}{*}{\#3} & source & Aber wenn Sie biologischen Evolution akzeptieren, bedenken Sie folgendes: \underline{ist es} nur über die Vergangenheit, oder geht es auch um die Zukunft?\\
& target & But if you accept biological evolution, consider this: Is it just about the past, or is it about the future?\\
& rating & $\sigma=0.48$, $\varnothing=1.12$\\ 
\midrule
\midrule
\multirow{3}{*}{\#4} & source & Finden Sie heraus, wie Sie überleben würden. \underline{Die meisten unserer Spieler haben die im Spiel gelernten Gewohnheiten beibehalten.}\\
& target & Find out how you would survive.\\
& rating & $\sigma=1.31$, $\varnothing=-0.79$\\
\midrule
\multirow{3}{*}{\#5} & source & Sie können das googlen, aber es ist keine Infektion des Rachens sondern der oberen Atemwege und verursacht den Verschluss der Atemwege.\\
 & target & You can \underline{googlen}, but it's not an infection of the \underline{rag}, but the upper respiratory \underline{pathway}, and it causes respiratory \underline{traction}.\\
& rating & $\sigma=1.31$, $\varnothing=-0.52$\\ 
\midrule
\multirow{3}{*}{\#6} & source & Nun, es scheint mir, dieses Thema wird, oder sollte wenigstens die interessanteste politische Debatte \underline{zum Verfolgen sein} über die nächsten paar Jahre.\\
& target & Well, it seems to me that this issue is going to be, or should be at least the most interesting political debate \underline{about} the next few years.\\
& rating & $\sigma=1.25$, $\varnothing=-0.93$\\
\bottomrule
\end{tabular}%
}
\caption{Items with lowest (top) and highest (bottom) deviation in 5-star ratings. Mean normalized rating and standard deviation are reported. Problematic parts of source and target are underlined, namely hallucinated or inadequate target words (\#1, \#5, \#6), over-literal translations (\#2), ungrammatical source (\#3, \#6) and omissions (\#4).}
\label{tab:5star-ex}
\end{table*}

\begin{table*}
\resizebox{\textwidth}{!}{
\begin{tabular}{ll|ll}
\toprule
\multirow{4}{*}{\#1} & source & Zu diesem Zeitpunkt haben wir mehrzellige Gemeinschaften, Gemeinschaften von vielen verschiedlichen Zellentypen, welche zusammen als einzelner Organismus fungieren.\\
& target1 & At this \underline{time} we have \underline{multi-tent} communities, communities of many different cell types, which act together as individual organism.\\
& target2 & At this point, we have multicellular communities, communities of many different cell types, which act together as individual organism.\\
& rating & $\sigma=0.0$, $\varnothing=1.0$\\
\midrule
\multirow{4}{*}{\#2} & source & Wir durchgehen dieselben Stufen, welche Mehrzellerorganismen durchgemacht haben – Die Abstraktion unserer Methoden, wie wir Daten festhalten, präsentieren, verarbeiten.\\
& target1 & We pass the same steps that \underline{have passed through} multi-cell organisms \underline{to process the abstraction} of our methods, how we record data.\\
& target2 & We go through the same steps that multicellular organisms have gone through -- the abstraction of our methods of \underline{holding} data, representing, processing.\\
& rating & $\sigma=0.0$, $\varnothing=1.0$\\
\midrule
\multirow{4}{*}{\#3} & source & Ich hielt meinen üblichen Vortrag, und danach sah sie mich an und sagte: "Mhmm. Mhmm. Mhmm."\\
& target1 & I \underline{thought} my usual talk, and then she looked at me and said:  \underline{mhmm}.\\
& target2 & I gave my usual talk, and then she looked at me and said, "mhmm. Mhmm. Mhmm."\\
& rating & $\sigma=0.0$, $\varnothing=1.0$\\
\midrule
\midrule
\multirow{4}{*}{\#4} & source & \underline{So in diesen Plänen}, wir hatten ungefähr 657 Plänen die den Menschen irgendetwas zwischen zwei bis 59 verschiedenen Fonds anboten. \\
& target1 & So in these plans, we had about 657 plans that offered \underline{the} people something between two to 59 different funds.\\
& target2 & So in these plans, we had about 657 plans that offered people anything between two to 59 different funds.\\
& rating & $\sigma=0.99$, $\varnothing=0.14$ \\
\midrule
\multirow{4}{*}{\#5} & source & Wir fingen dann an, über Musik zu sprechen, angefangen von Bach über Beethoven, Brahms, Bruckner und all die anderen Bs, von Bartók bis hin zu Esa-Pekka Salonen.\\ 
& target1 & We then began to talk about music, \underline{starting from} \underline{b}ach on Beethoven, Brahms, Bruckner and all the other \underline{b}s, from Bartók to \underline{e}sa-pekka \underline{salons}.\\
& target2 & We started talking about music from \underline{b}ach, Beethoven, Brahms, Bruckner and all the other \underline{b}s, from Bart\underline{o}k to \underline{e}sa-pekka \underline{salons}.\\
& rating & $\sigma=0.99$, $\varnothing=-0.14$\\
\midrule
\multirow{4}{*}{\#6} & source & Heinrich muss auf all dies warten, nicht weil er tatsächlich ein anderes biologische Alter hat, nur aufgrund des Zeitpunktes seiner Geburt.\\
& target1 & Heinrich has to wait for all of this, not because \underline{he's actually having} another biological age, just because of the time of his birth.\\
& target2 & Heinrich must wait for all this, not because he actually has another biological age, only due to the time of his birth.\\
& rating & $\sigma=0.99$, $\varnothing=-0.14$\\
\bottomrule
\end{tabular}%
}
\caption{Items with lowest (top) and highest (bottom) deviation in pairwise ratings. Preferences of target1 are treated as "-1"-ratings, preferences of target2 as "1", no preference as "0", so that a mean ratings of e.g. -0.14 expresses a slight preference of target1. Problematic parts of source and targets are underlined, namely hallucinated or inadequate target words (\#1, \#2, \#3, \#4), incorrect target logic (\#2), omissions (\#3), ungrammatical source (\#4), capitalization (\#5), over-literal translations (\#5, \#6).}
\label{tab:pw-ex}
\end{table*}

\section{Reward Estimation}
\subsection{Auxiliary Data for Reward Estimation} \label{app:re}
In order to augment the small collection of 1,000 rated translations, we leverage the available out-of-domain bitext as auxiliary training data: 10k source sentences of WMT (out-of-domain) are translated by the out-of-domain model. Translations from 9 beam search ranks are compared to their references to compute sBLEU rewards. This auxiliary data hence provides 90k out-of-domain training samples with sBLEU reward. For pairwise rewards, sBLEU scores for two translations for the same source are compared. Each mini-batch during training is sampled from the auxiliary data with probability $p_{aux}$, from the original training data with probability $1-p_{aux}$. Adding this auxiliary data as a regularization through multi-task learning prevents the model from overfitting to the small set of human ratings. In our experiments, $p_{aux}=0.8$ worked best.

\subsection{Reward Estimation Architecture} \label{app:nn}

\begin{figure}
\includegraphics[width=\columnwidth]{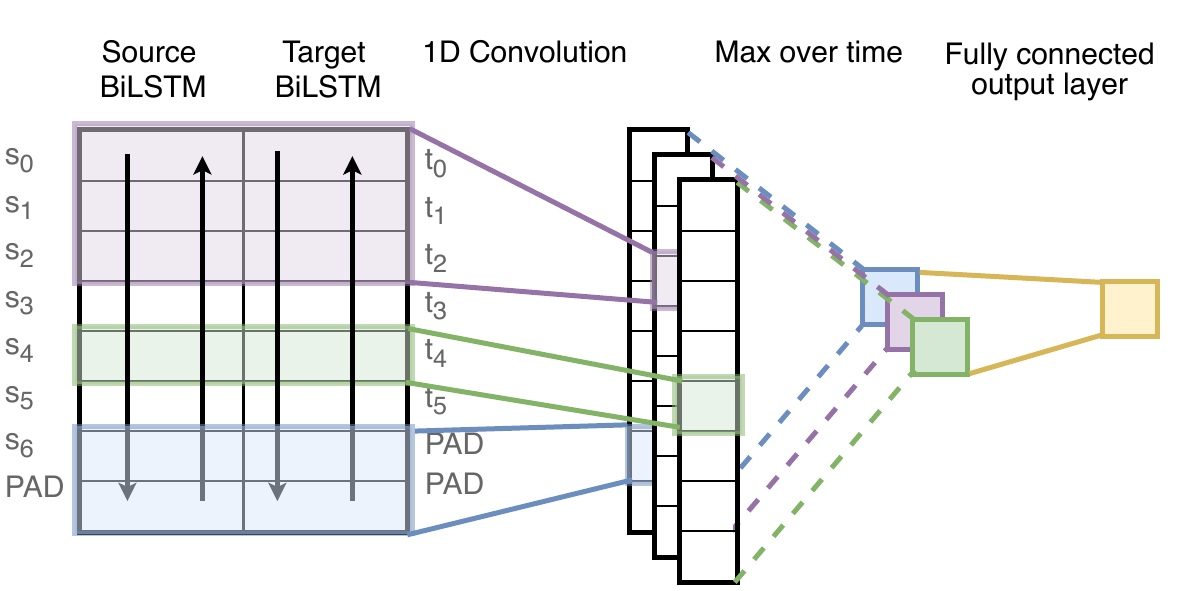}
\caption{Reward estimation architecture: Source and target biLSTM outputs over subword embeddings are concatenated for each position, followed by a convolutional layer with several filters (here one each for sizes 1 to 3), a max-over-time pooling and a fully connected output layer with a single leaky ReLU output unit.}
\label{fig:re-architecture}
\end{figure}

Input source and target sequence are split into the BPE subwords used for NMT training, padded up to a maximum length of 100 tokens, and represented as 500-dimensional subword embeddings. Subword embeddings are pre-trained on the WMT bitext with \texttt{word2vec} \cite{MikolovETAL:13}, normalized to unit length and held constant during further training. Additional 10-dimensional BPE-feature embeddings are appended to the subword embeddings, where a binary indicator encodes whether each subword contains the subword prefix marker "@@". BPE-prefix features are useful information for the model since bad translations can arise from ``illegal'' compositions of subword tokens. The embeddings are then fed to a ssource-side and a target-side bidirectional LSTM (biLSTM) \cite{HochreiterSchmidhuber:97}, respectively. The biLSTM outputs are concatenated for each time step and fed to a 1-D convolutional layer with 50 filters each for filter sizes from 2 to 15. The convolution is followed by max-over-time pooling, producing 700 input features for a fully-connected output layer with leaky ReLU \cite{MaasETAL:13} activation function. Dropout \cite{SrivastavaETAL:14} with $p=0.5$ is applied before the final layer. This architecture, depicted in Figure \ref{fig:re-architecture}, can be seen as a biLSTM-enhanced bilingual extension to the convolutional model for sentence classification proposed by \citet{Kim:14}. 
 
\section{NMT}
\subsection{NMT Hyperparameters}\label{app:hyper}
The NMT has a bidirectional encoder and a single-layer decoder with 1,024 GRUs each, and subword embeddings of size 500 for a shared vocabulary of subwords obtained from 30k byte-pair merges \cite{SennrichETAL:16}.
 Maximum input and output sequence length are set to 60. 
For the MLE training of the out-of-domain model, we optimize the parameters with Adam ($\alpha=10^{-4}$, $\beta_1=0.9$, $\beta_2=0.999$, $\epsilon=10^{-8}$) \cite{KingmaBa:14}. For further in-domain tuning (supervised, OPL and RL), $\alpha$ is reduced to $10^{-5}$. To prevent the models from overfitting, dropout with probability 0.2 \cite{SrivastavaETAL:14} and l2-regularization with weight $10^{-8}$ are applied during training. The gradient is clipped to its norm when its norm exceeds 1.0 \cite{PascanuETAL:13}. Early stopping points are determined on the respective development sets. For model selection we use greedy decoding, for test set evaluation beam search with a beam of width 10. For MLE and OPL models, mini-batches of size 60 are used. For the RL models, we reduce the batch size to 20 to fit $k=5$ samples for each source into memory. The temperature is furthermore set to $\tau=0.5$. We found that learning rate and temperature were the most critical hyperparameters and tuned both on the development set.

\end{document}